 \documentclass[preprint,12pt]{elsarticle}

\usepackage{epstopdf}

\usepackage{amssymb}
\usepackage{amsthm}

\usepackage{algpseudocode}
\usepackage{algorithm}

 \usepackage{listings}
 \usepackage{courier}
\usepackage[nodots]{numcompress}

\usepackage{lineno}

\begin{document}

\begin{frontmatter}



\title{Veni Vidi Vici, A Three-Phase Scenario For Parameter Space Analysis in Image Analysis and Visualization}


\author[label1]{Mohammed A. El-Dosuky}

\address[label1]{Computer sciences Department, Faculty of Computers and Information, P.D.Box 35516, Mansoura University, Egypt}

\begin{abstract}
Automatic analysis of the enormous sets of images is a critical task in life sciences. This faces many challenges such as: algorithms are highly parameterized, significant human input is intertwined, and lacking a standard meta-visualization approach. This paper proposes an alternative iterative approach for optimizing input parameters, saving time by minimizing the user involvement, and allowing for understanding the workflow of algorithms and discovering new ones. The main focus is on developing an interactive visualization technique that enables users to analyze the relationships between sampled input parameters and corresponding output.  This technique is implemented as a prototype called Veni Vidi Vici, or "I came, I saw, I conquered." This strategy is inspired by the mathematical formulas of numbering computable functions and is developed atop ImageJ, a scientific image processing program. A case study is presented to investigate the proposed framework. Finally, the paper explores some potential future issues in the application of the proposed approach in parameter space analysis in visualization.
\end{abstract}


\end{frontmatter}


\section{Introduction}
Visual analysis is a vital tool in life sciences. Scientists can automate experiments and capture large result sets of digital images \cite{Scientists10}. Automatic analysis of the enormous sets of biomedical images is a must.
Algorithms and techniques of image analysis are highly parameterized and significant human input is intertwined to optimize parameter settings \cite{Ruddle11}.
It is common to utilize machine-learning algorithms in minimizing user overhead. However, users are asked to determine objects of interest in some input images, then the algorithms create classifiers to recognize similar objects in other images. This approach highly relies on the experience of domain experts and it requires a great amount of object labeling to be done by hand.

Also, lacking a standard meta-visualization\cite{meta11} approach is a challenge. High-level description of a visualization algorithm can not be gained by debugging, forcing developers of the algorithm to manually create high-level illustrations to understand the workflow of the algorithm.

This paper proposes an alternative iterative approach for optimizing input parameters, saving time by minimizing the user involvement, and allowing for understanding the workflow of algorithms and discovering new ones. The main focus is on developing an interactive visualization technique that enables users to analyze the relationships between sampled input parameters and corresponding output.  This technique is implemented in a prototype called Veni Vidi Vici , following the famous Latin sentence claimed to be said by Julius Caesar \cite{venividivici} . It translates as "I came, I saw, I conquered." This strategy is inspired by the mathematical formulas of numbering computable functions \cite{Cutland80} and is developed atop ImageJ \cite{ImageJ}, a scientific biomedical image analysis program.

Section 2 summarizes related work, before surveying ImageJ in Section 3.
Proposed work is presented in Section 4. This includes explaining Veni Vidi Vici strategy  (Section 4.1), presenting the underpinning mathematical foundations (Section 4.2), and implementing the framework as a custom plug-in for ImageJ  (Section 4.3). Evaluation of the proposed approach with case studies are shown in Section 5. Finally, presenting conclusion and outlining future work in Section 6.

\section{Related Work}
Many visualization techniques are proposed to facilitate investigating the relationships between parameters and outcomes. Linked charts are among the first attempts to use interactive visualization to parameter optimization \cite{Tweedie95}. The mapping $p_{1}, ..., p_{n}$ $\to$ $o_{1}, ...,o_{m}$, where $p_{1}, ..., p_{n}$ are the parameters and $o_{1}, ...,o_{m}$ are the outcomes, can be  represented as n + m interactive histograms. This approach is then extended to prosection matrices \cite{Tweedie98}. An alternative approach is proposed for visualizing the parameter search process as a directed graph \cite{Ma99}.

Exploration graphs for history management are used in VisTrails visualization management system \cite{VisTrails05}. The exploration is still linear and time consuming. This can be partially improved by allowing users to write code snippets to perform iterations over parameter values \cite{VisTrails06}.

Many techniques are proposed for image-based analysis and visualization as a structured gallery \cite{JankunKelly0}. An innovative technique is proposed based on preset based interaction with high dimensional parameter spaces \cite{Preset03}. A preset is a reference point in parameter space used to compute the outcome by calculating the linear combination of the inverse distances from the preset to the outcome.

Many successful analysis and visualization systems are available such as IRIS Explorer\cite{IRIS95}. CellProfiler \cite{CellProfiler06} is an image analysis software for identifying and quantifying cell phenotypes. VolumeShop \cite{VolumeShop05} is an interactive system for direct volume illustration.

Vismon \cite{Maryam11} is a data analysis visual tool for fisheries to enable decision makers to quickly narrow down all possible management options to an agreeable small set. A risk assessment framework for chum salmon is developed to evaluate different management policies \cite{CPZ09}. A detailed trade-off among the few chosen management options can be performed \cite{Maryam12}.

Most of previous systems are domain-specific. However, some are for a wide range of applicability. Tuner \cite{Tuner11} is an application for finding optimal parameter settings for complex algorithms in high-dimensional parameter space. It supports sampling, re-sampling, and visually seeing the stability of certain parameter settings with respect to the objective measures.

For more wisdom on new trends in visualization pipeline, algorithmic stability, and the exploration of parameters, readers are encouraged to check IEEE VisWeek 2012 tutorial on "Uncertainty and Parameter Space Analysis in Visualization" \cite{VisWeek12}.
\section{ImageJ}
ImageJ \cite{ImageJ} is a free image processing program developed at the National Institutes of Health.  Originally, it is designed to be used by biomedical researchers working with microscope images and videos.

Being based on Java which supports algorithmic digital image processing\cite{Java7}, ImageJ is designed with an open architecture that provides extensibility via Java plugins and recordable macros\cite{ImageJ2}. These plugins provide custom solution for many analysis and visualization problems such as:
\begin{itemize}
  \item three-dimensional live-cell imaging\cite{ImageJ3},
  \item to radiological image processing\cite{ImageJ4},
  \item multiple imaging system data comparisons\cite{ImageJ5},
  \item automated hematology systems\cite{ImageJ6}, and
  \item Visualizing multidimensional biological image data\cite{ImageJ7}.
\end{itemize}
The following four subsections show a comprehensive list of plugins.
\subsection{Analysis}
\begin{itemize}
  \item Autocorrelation, 	
  \item MRI t2 calculations,
  \item Line Analyzer,	
  \item Image Correlator , 	
  \item Particle Remover,
  \item Circularity ,	
  \item Modulation Transfer Function, 	
  \item Specify ROI ,
  \item Specify Line Selection,	
  \item 16-bit Histogram ,
  \item Draw line or point grids ,
  \item Moment Calculator ,
  \item Batch Statistics ,	
  \item Cell Counter ,
  \item Oval Profile Plot ,
  \item Color Comparison ,	
  \item Radial Profile Plot ,
  \item Microscope Scale ,
  \item MRI Analysis Calculator ,
  \item Sync Measure 3D , 	
  \item Hough Circles ,
  \item Convex Hull, Circularity, Roundness ,
  \item Fractal Dimension and Lacunarity ,
  \item Measure And Label ,
  \item Colocalization ,
  \item Granulometry ,
  \item Texture Analysis ,
  \item Named Measurements ,
  \item Cell Outliner ,
  \item Grid Cycloid Arc ,
  \item RGB Profiler ,
  \item Colocalization Finder ,
  \item Spectrum Extractor ,
  \item Contact Angle ,
  \item RG2B Colocalization ,
  \item Color Profiler ,
  \item Hull and Circle ,
  \item MR Urography ,	
  \item Template Matching ,
  \item Extract IMT from ultrasound images ,
  \item ITCN (Image-based Tool for Counting Nuclei) ,
  \item Multi Cell Outliner ,
  \item FRETcalc - FRET by acceptor photobleaching ,
  \item JACoP (Just Another Colocalization Plugin),
  \item FRET and Colocalization Analyzer ,
  \item CASA (Computer Assisted Sperm Analyzer) ,
  \item Radial Profile Plot Extended ,
  \item Concentric Circles (non-destructive overlay),	
  \item Azimuthal Average ,
  \item Slanted Edge Modulation Transfer Function,
  \item Calculate 3D Noise ,
  \item FWHM (analyze photon detector pinhole images),	
  \item SSIM index (calculate structural similarity),
  \item Image Moments (image moments of n-th rank) ,
  \item MS SSIM index (multi-scale structural similarity),	
  \item Colony Counter (count colonies in agar plates),
  \item Levan (chromosome morphology) ,
  \item EXTRAX (electron diffraction intensity extraction), 	
  \item Fractal Surface Measurement ,
  \item Foci Picker3D (finds local maxima in 2D and 3D images),	
  \item Diameter (measures the diameter of a blood vessels),	
  \item Graph Demo (creates particle adjacency lists) ,
  \item Asymmetry Analysis (HR-TEM image conditions) ,	
  \item 2D NMR Analysis (integrates peaks in 2D NMR spectra) ,	
  \item MetaData and Intracellular Calcium Line Scan Analysis,
  \item GHT (General Hough Transformation object recognition),
  \item IntraCell (nanoparticle colocalization within cells),
  \item Lemos Asymmetry Analysis (dental panoramic radiographs),
  \item Merz Grid Macro (semicircular lines and points in overlay),
  \item Stress Granule Counter (counts SGs in eucaryotic cells),
  \item Vamp 2D and 3D (isolate puncta in 2D and 3D images) ,
  \item Sampling Window (unbiased sampling window) ,
  \item Map Bone Microstructure (histomorphometry parameters),
  \item Results and Text ,
  \item Comment Writer
\end{itemize}

\subsection{Filters}
\begin{itemize}
     \item Real Convolver
     \item Fast Fourier Transform (FFT)
     \item LoG Filtering
     \item Background Subtraction and Normalization
     \item Contrast Enhancer
     \item Background Correction
     \item Byte Swapper
     \item Discrete Cosine Transform (DCT)
     \item FFT Filter
     \item FFTJ and DeconvolutionJ
     \item Unpack 12-bit Images
     \item De-interlace
     \item 2D Gaussian Filter
     \item Kalman Filter
     \item Dual-Energy Algorithm
     \item Anisotropic Diffusion  (edge-preserving noise reduction)
     \item Grayscale Morphology
     \item 2D Hybrid Median Filter
     \item 3D Hybrid Median Filter
     \item Spectral Unmixing
     \item Haar Wavelet Filter and Adaptive Median Filter
     \item 'A trous' Wavelet Filter
     \item Kuwahara Filter
     \item Granulometric Filtering
     \item Windowed-Sinc Filter  (low pass time series filter)
     \item Anisotropic Diffusion 2D  (edge-preserving noise reduction)
     \item Auto Gamma  for gamma correction
     \item Linearize Gel Data
     \item Radon Transform  (back projection, sinogram)
     \item Correct X Shift of Confocal Images
     \item Multi Otsu Threshold
     \item Spectral Unmixing of Bioluminescence Signals
     \item Lipschitz Filter
     \item Float Morphology  (erode, dilate, open, close)
     \item X Shifter  (correct pixel mismatch of confocal images)
     \item Sigma Filter  (edge-preserving noise reduction)
     \item Rolling Ball Background Subtraction
     \item Mean Shift Filter  (edge-preserving smoothing)
     \item Accurate Gaussian Blur
     \item Add Poisson Noise
     \item CLAHE  (Contrast Limited Adaptive Histogram Equalization)
     \item Floyd Steinberg Dithering
     \item Polar Transformer  (corrects radial and angular distortions)
     \item Gaussian Blur 3D
     \item Image Rotator  (rotates image around ROI center of mass)
     \item Mexican Hat  (2D Laplacian of Gaussian)
\end{itemize}
\subsection{Segmentation}
\begin{itemize}
  \item Mixture Modeling Thresholding ,
  \item Otsu Thresholding ,
  \item Watershed Segmentation Maximum Entropy Thresholding ,	
  \item MultiThresholder (IsoData, MaxEntropy, Otsu, etc), 	
  \item Multi Otsu Threshold ,
  \item SIOX (Simple Interactive Object Extraction),	
  \item RATS (Robust Automatic Threshold Selection) ,
  \item Densitometry ,
  \item Blob Labeler (labels connected blobs of pixels)	
\end{itemize}

\subsection{Collections}
\begin{itemize}
  \item UCSD Confocal Microscopy Plugins , 	
  \item MBF ImageJ for Microscopy Collection
\end{itemize}

\section{Proposed Framework}
\subsection{Veni Vidi Vici}
Let us discuss the general idea behind the proposed new iterative approach for optimizing input parameters and allowing for understanding the workflow of algorithms and discovering new ones. The main focus is on developing an interactive visualization technique that enables users to analyze the relationships between sampled input parameters and corresponding output.  This technique is implemented in a prototype called Veni Vidi Vici , following the famous Latin sentence claimed to be said by Julius Caesar \cite{venividivici}. It translates as "I came, I saw, I conquered."

It falls into three main parts:
\begin{itemize}
  \item Veni: corresponds to analysis and user interaction,
  \item Vidi: corresponds to visualization, and
  \item Vici: corresponds to segmentation or similar tasks.
\end{itemize}
Figure 1 shows the schematic diagram of the proposed strategy.

\begin{figure}[ht!]
\centering
\includegraphics[bb=0 0 381 161]{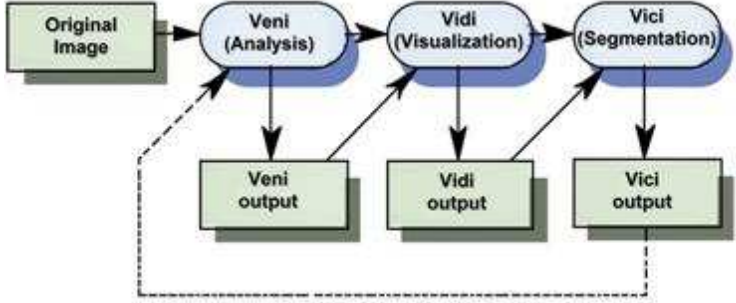}
\caption{Proposed Framwork}
\label{framwork}
\end{figure}

Note that rectangles represent files and rounded rectangles represent processes.
Veni takes the original image and the outcome of previous runs as input. The dashed line is not presented in the first run. User interaction is minimized to Veni. Vidi is sandwiched between the other two processes. Vici corresponds to a desired task to perform on the image such as segmentation or counting objects. The final outcome is then fed-back to Veni.
These three processes are optional ,i.e. any one can be skipped, with at least one applied.
\subsection{Mathematical Foundations}
Recall that the over all process involves a set of n parameters: $p_{1}, ..., p_{n}$.
Assume that these n parameters are divided among the three processes of Veni, Vidi, and Vici.
Assume that $a, b,$ and $c$ are the parameter shares of Veni, Vidi, and Vici respectively.
So, $n = a + b + c$. Veni, Vidi, and Vici are descried mathematically in table 1.

\begin{table}[ht]
\caption{Parameter share}
\begin{center}
    \begin{tabular}{ | l | p{5cm} |}
    \hline
    Operation & Summary \\ \hline
    Veni(g, $x_{1}, ..., x_{a}$) & Apply Veni on images g, setting its a parameters to x's values\\ \hline
    Vidi(g, $y_{1}, ..., y_{b}$) & Apply Vidi on images g, setting its b parameters to y's values\\ \hline
    Vici(g, $z_{1}, ..., z_{c}$) & Apply Vici on images g, setting its c parameters to z's values\\ \hline
    \end{tabular}
\end{center}
\end{table}

The proposed strategy for exploring parameter space is inspired by the mathematical formulas of numbering computable functions \cite{Cutland80}. Let us define some functions that allow us to map any set of  Veni, Vidi, and Vici to
a unique code, or G\"{o}del number \cite{godel}.  	
\\
\\
\textbf{Mapping ordered pairs to N}
 	
A function $\pi$ such that 	
$ \pi(x,y) = 2^x(2y + 1) - 1 $
 maps $N^2 \to N$. That is, $\pi$ maps the ordered pair (x,y) to a single number $2^x(2y + 1) - 1$. 	

\begin{figure}[ht!]
\centering
\includegraphics[bb=0 0 381 161]{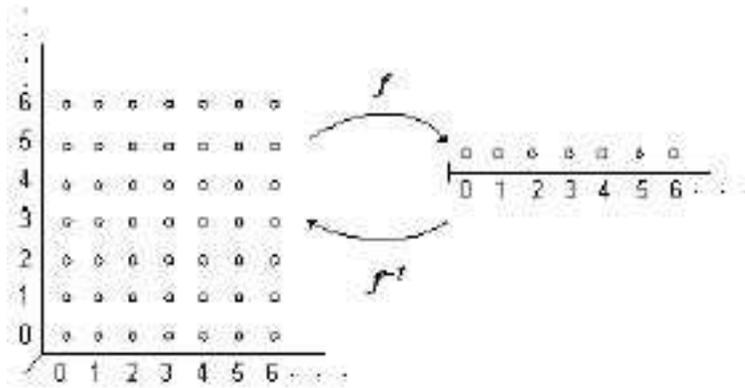}
\caption{A graphical representation of $\pi$}
\label{pi}
\end{figure}

To see how this mapping is done, note that $\pi = 2^x(2y+ 1) - 1$ is a number such that $\pi$ has x factors of 2 and a remaining odd number 2y + 1.
As an example, let $\pi$ = 103935. You can factories $\pi$ + 1 into 9 factors of 2 and you are left over  with $203 = 101(2) - 1$. Hence, $\pi + 2 = 2^9(2(101) + 1)$ yielding x = 9 and y = 101. Since every number has a different factorization into powers of two, $\pi$	
is a bijective function mapping. The inverse of $\pi$ is defined such that $\pi_1$(z) 	
returns the exponent of $p_1$ = 2 in the prime factorization of z, and $\pi_2$(z) returns the exponent of $p_2$ = 3.
\\
\\
\textbf{Mapping Ordered Triples to N}

A function 	$ \zeta : N^+ \times N^+ \times N^+ \to N $	 maps an ordered set of three natural numbers to a single number. To define this mapping, we can use the function $\pi$, and write $ \zeta (m,n,q) = \pi(\pi(m-1,n-1),q -1)$ which maps the first parameters m and n to one number using $\pi$. The resulting value and the third number q is then mapped to one number again using $\pi$. Hence, the final result is one natural number. The purpose in subtracting 1 from the parameters m,n and q is to include zero in the mapping.
\\
\\
\textbf{Mapping a Finite Sequence $N^k$ to N}

The concept of converting bases can be used to devise a bijection from a finite sequence of numbers to one natural number.

The binary representation is simple yet powerful. A decimal number can be converted into a binary representation by repeatedly subtracting the next largest possible power of two from the previous remainder, until no other powers exist.

Consider the conversion of 44 decimal to binary: $44 = 2^5 + 12 = 2^5 + 2^3 + 4 = 2^5 + 2^3 + 2^2$. By putting 1 for every existing power of 2, the binary representation of 44 is 101100.
Let us define a function mapping from a finite sequence of numbers to one natural number. Let the function $\tau(a_1,a_2,...,a_k ) \to N$ be defined as $\tau(a_1,a_2,...,a_k ) = 2^{a_1} + 2^{a_1 + a_2 + 1} + 2^{a_1 + a_2 + a_3 + 2} + ··· +2^{a_1 + a_2 + a_3 + ... + a_k + (k-1)} - 1$ 	
As an example, let us perform the calculations for $\tau (5, 3, 2):$
$\tau (5, 3, 2) = 2^5  +2^{5+3+1} +2^{5+3+2+2} - 1 	$

\subsection{Implementation}
Let us propose the following algorithm.
\begin{algorithm}[H]
  \begin{algorithmic}[1]
    \Statex \Comment{ Load Default settings }
    \State $settings \leftarrow Load\_Default\_Settings()$
    \Statex \Comment{ Parameter share is set}
    \State $a \leftarrow Input\_Integer()$
    \State $b \leftarrow Input\_Integer()$
    \State $c \leftarrow Input\_Integer()$
    \Statex \Comment{ Exploration range is set}
    \State $range \leftarrow Input\_Integer()$
    \Statex \Comment{ Initially, $cur$ contains input images}
    \State $cur \leftarrow Get\_Input\_Images()$
    \While{ ( TRUE ) }
    \State $[cur, Veni_{code}]\leftarrow Veni(cur, settings[1..a])$
    \State $Save\_Files(cur)$
    \State $[cur, Vidi_{code}]\leftarrow Vidi(cur, settings[a+1..a+b])$
    \State $Save\_Files(cur)$
    \State $[cur, Vici_{code}] \leftarrow Vici(cur, settings[a+b+1..a+b+c])$
    \State $Save\_Files(cur)$
    \Statex \Comment{ Encode the Veni Vidi Vici operations}
    \State $code \leftarrow Encode(Veni_{code}, Vidi_{code}, Vici_{code})$
    \Statex\Comment{ enumerate outcomes, one by one}
    \For  {$i= code - range / 2$ to $code + range / 2$}
    \State        $temp\_settings[i] \leftarrow Decode(i)$
    \State        $Image\_View(i, cur, temp\_settings[i])$
    \EndFor
    \Statex \Comment{ User Interaction}
    \State $Timed\_Pause()$
    \State $selection \leftarrow Input\_Integer()$
    \Statex\Comment{ Take the current settings.}
    \If{  ( $selection = NIL$) }
    \State $Exit While$
    \Else
    \State $settings \leftarrow temp\_settings[selection]$
    \EndIf
    \EndWhile
  \end{algorithmic}
  \caption{Veni Vidi Vici Algorithm}
\end{algorithm}

The algorithm starts with loading default settings from setting file containing the values on $n$ parameters. The parameters are divided among Veni Vidi Vici operations.
The algorithm also asks for the exploration range. The smaller it is, the fewer possibilities the algorithm examines.
Then the algorithm load an image or a set of images, and perform successive Veni Vidi Vici operations. Each operation emits a set of files, and has a code calculated using the underpinning mathematical foundations presented in Section 4.2.
The total code of Veni Vidi Vici operations is then calculated in the same way.

The total code allows for enumerating outcomes, one by one along the exploration range.
Each outcome has a corresponding decoded settings. In 17\textsuperscript{th} line, user interaction is required to determine which settings suits his or her needs.

The framework can be implemented in many ways such as a recordable macro or a plug-in as shown in figure 3.

\begin{figure}[ht!]
\centering
\includegraphics[bb=0 0 310 361]{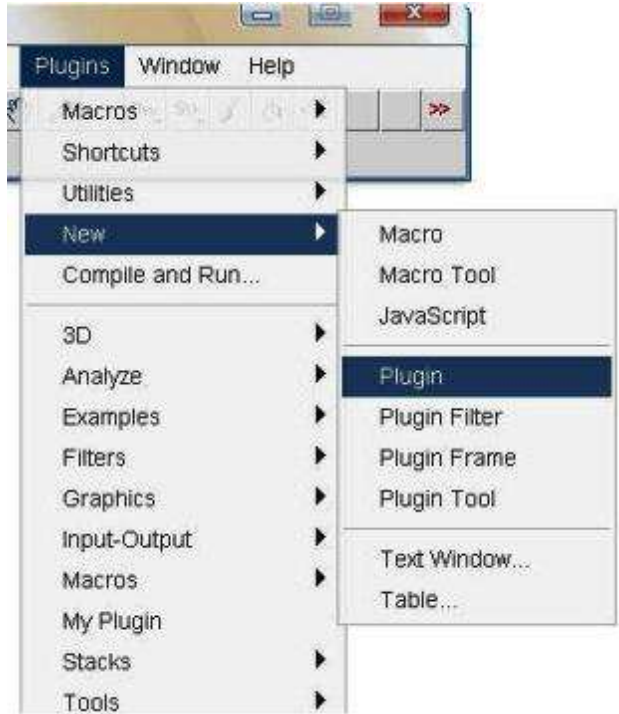}
\caption{Creating a new plug-in}
\label{plugin}
\end{figure}

An editor pop ups to modify the run method. After compiling the plug-in, the created plug-in is shown in figure 4. Note that the name of the plug-in must contains underscores to be automatically loaded when ImageJ starts.

\begin{figure}[ht!]
\centering
\includegraphics[bb=0 0 280 375]{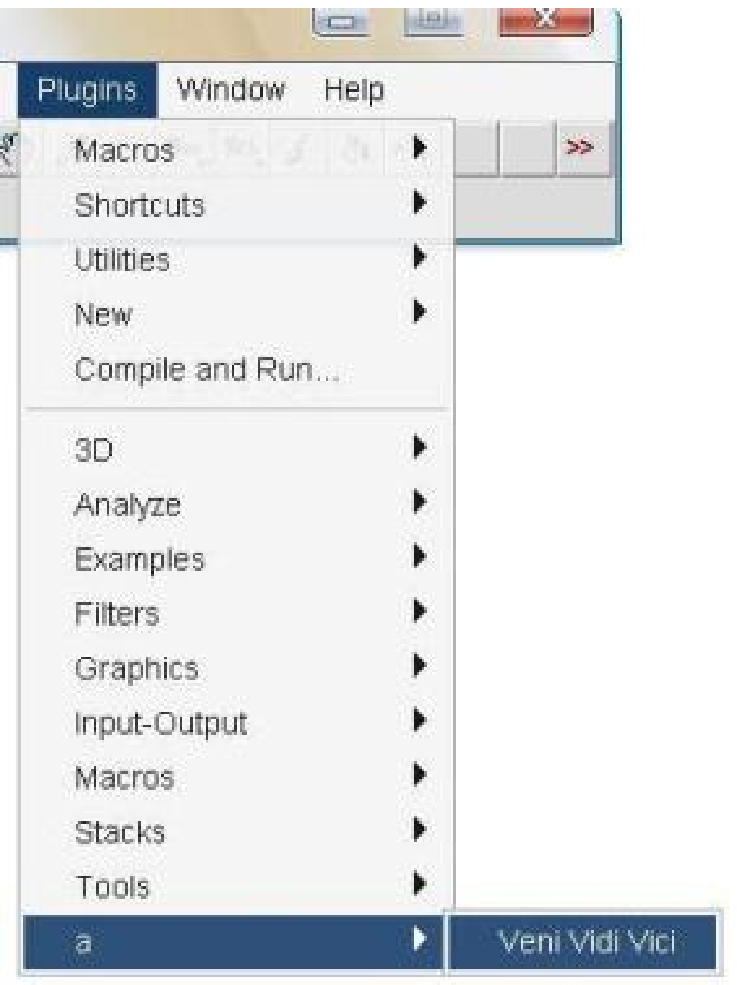}
\caption{The created plug-in}
\label{vvv}
\end{figure}

\section{Evaluation}
To load an image, just drag-drop it on ImageJ.
By clicking the menu item for the plug-in, it starts working.

Many trails were done to test the proposed framework.
This was done on images provided on ImageJ site, shown in figure 5.

\begin{figure}[ht!]
\centering
\includegraphics[bb=0 0 335 245]{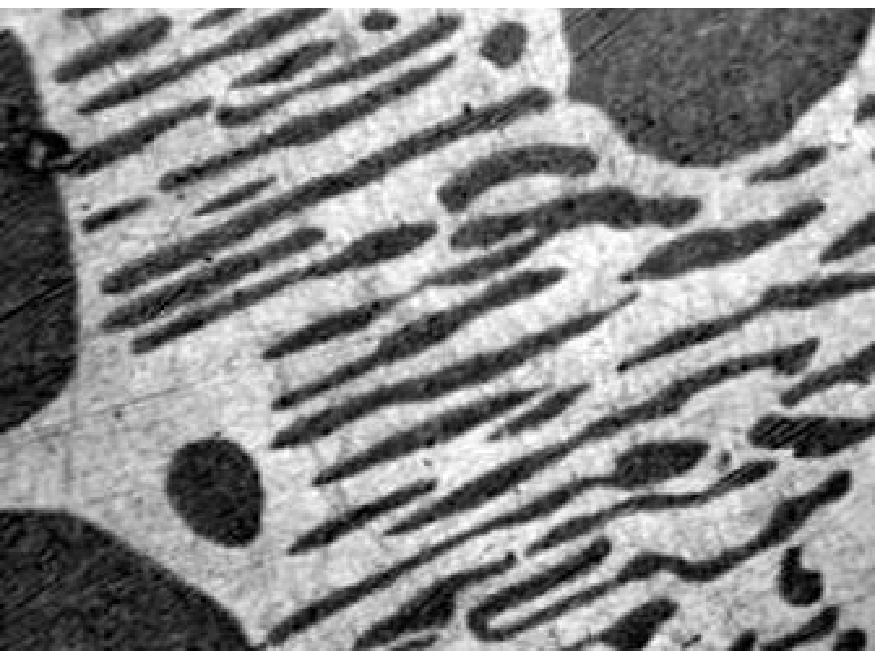}
\caption{The original image}
\label{o}
\end{figure}

Many Veni operations can be done on the loaded image such as analysis plugins listed in section 3.
Figures 6,7,8 shows the application of edge detection, plot profile, and surface plot.

\begin{figure}[ht!]
\centering
\includegraphics[bb=0 0 335 245]{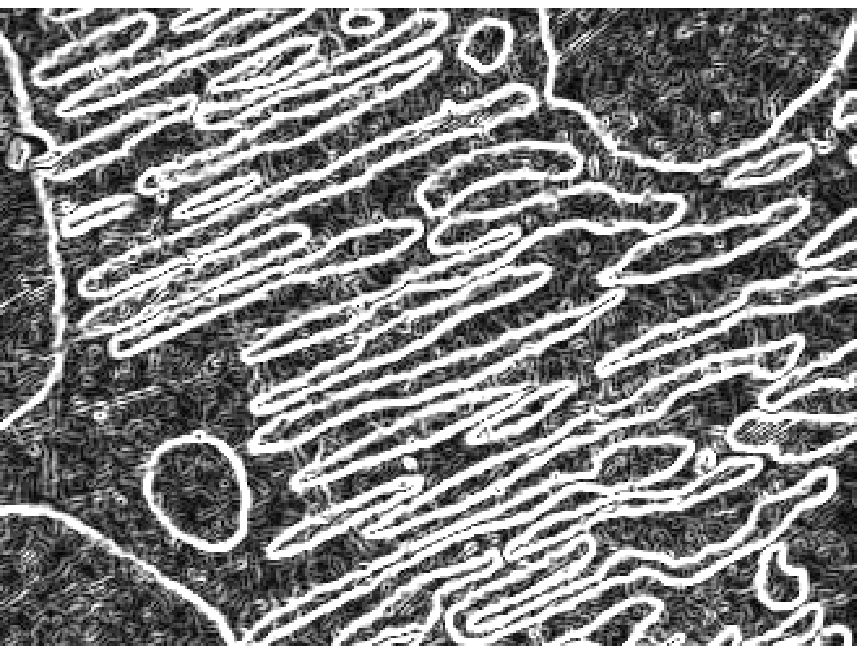}
\caption{edge detection}
\label{e}
\end{figure}

\begin{figure}[ht!]
\centering
\includegraphics[bb=0 0 335 245]{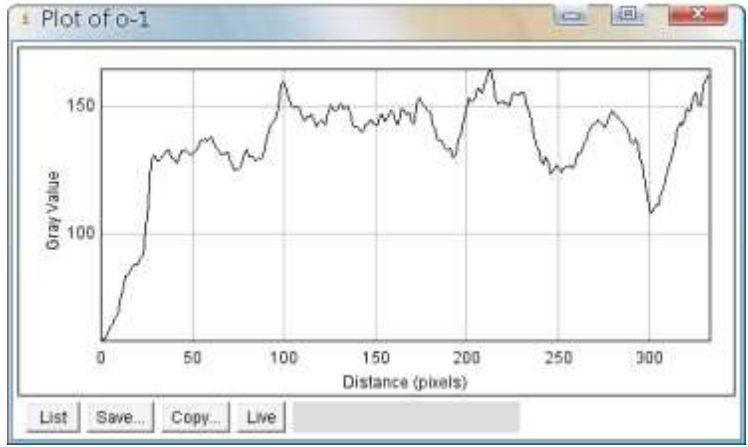}
\caption{plot profile}
\label{pp}
\end{figure}

\begin{figure}[ht!]
\centering
\includegraphics[bb=0 0 335 245]{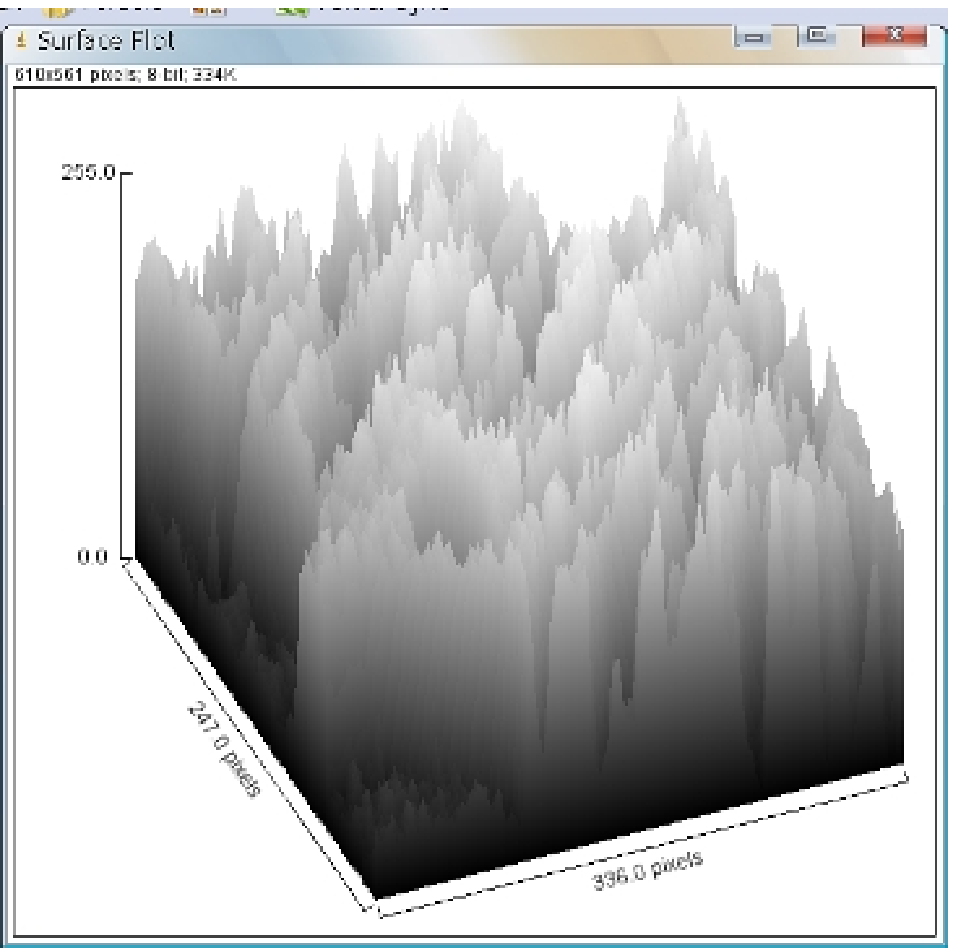}
\caption{surface plot}
\label{sp}
\end{figure}

Many Vici operations can be done such as segmentation plugins listed in section 3.
Figure 9 shows the application of Otsu thresholding technique \cite{otsu79}. This technique divides the histogram in two classes and then the inter-class variance is minimized. This plugin outputs a thresholded image with the selected threshold.

\begin{figure}[ht!]
\centering
\includegraphics[bb=0 0 335 245]{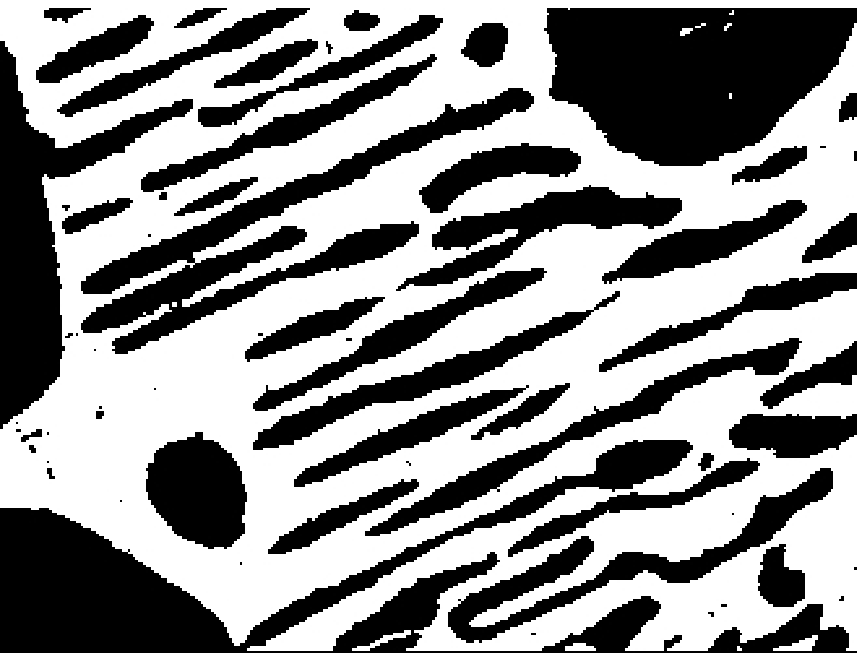}
\caption{Otsu thresholding}
\label{ot}
\end{figure}

\section{Conclusion and Future Work}
This paper proposes an alternative iterative approach to optimize input parameters and save time by minimizing the user involvement. This strategy is developed as custom plug-in in ImageJ.

The main focus was on developing an interactive visualization technique that enables users to analyze the relationships between sampled input parameters and corresponding output.  This technique is implemented in a prototype called Veni Vidi Vici.
It provides users with a visual overview of parameters and their sampled values. To find optimal parameter settings, users can select best possible configuration.

Not only the proposed framework facilitates better algorithm understanding, but also it can be used to explore the parameter space.

Future work have many dimension.
One dimension is to consider the application of the proposed framework on temporal data such as feature point tracking and trajectory analysis for video imaging \cite{mosaic}.

Another dimension is extending the theoretical foundation of the proposed framework to be able to answer some questions regarding the parameters of Veni, Vidi, and Vici. Questions such as : How does a slight parameter change modify the result? How stable is a parameter? In which range is a parameter stable?

Another dimension is to enhance the accuracy of the framework by incorporating transfer learning \cite{Sinno}, to be able to generalize what learnt from one case to another. In many machine learning, there is an assumption that the training and future data must be in the same feature space and have the same distribution. However, in many real-world applications, this assumption may not hold. Knowledge transfer, if done successfully, would greatly improve the performance of learning by avoiding much expensive data-labeling efforts.





\bibliographystyle{model3-num-names}



\end{document}